\documentclass[10pt,twocolumn,letterpaper]{article}

\usepackage[pagenumbers]{cvpr} 

\usepackage{graphicx}
\usepackage{amsmath}
\usepackage{amssymb}
\usepackage{booktabs}
\usepackage{multirow}

\usepackage[dvipsnames]{xcolor}

\usepackage{colortbl}

\usepackage[pagebackref,breaklinks,colorlinks]{hyperref}

\usepackage[capitalize]{cleveref}
\crefname{section}{Sec.}{Secs.}
\Crefname{section}{Section}{Sections}
\Crefname{table}{Table}{Tables}
\crefname{table}{Tab.}{Tabs.}

\begin{document}

\title{ConQueR: Query Contrast Voxel-DETR for 3D Object Detection}

\author{Benjin Zhu\textsuperscript{\rm 1} \quad
        Zhe Wang\textsuperscript{\rm 1} \quad
        Shaoshuai Shi\textsuperscript{\rm 2} \quad
        Hang Xu\textsuperscript{\rm 3} \quad
        Lanqing Hong\textsuperscript{\rm 3} \quad
        Hongsheng Li\textsuperscript{\rm 1} \\
\textsuperscript{\rm 1}Multimedia Laboratory, The Chinese University of Hong Kong \\
\textsuperscript{\rm 2}Max Planck Institute for Informatics \\
\textsuperscript{\rm 3}Huawei Noah's Ark Lab \\
{\tt\small \{benjinzhu@link,hsli@ee\}.cuhk.edu.hk}
}

\maketitle

\begin{abstract}
Although DETR-based 3D detectors can simplify the detection pipeline and achieve \textbf{direct} sparse predictions, their performance still lags behind dense detectors with post-processing for 3D object detection from point clouds. DETRs usually adopt a larger number of queries than GTs (\emph{e.g.}, 300 queries v.s.~$\sim$40 objects in Waymo) in a scene, which inevitably incur many false positives during inference. In this paper, we propose a simple yet effective sparse 3D detector, named \textbf{Que}ry \textbf{Con}trast Voxel-DET\textbf{R} (\textbf{ConQueR}), to eliminate the challenging false positives, and achieve more accurate and sparser predictions. We observe that most false positives are highly overlapping in local regions, caused by the lack of explicit supervision to discriminate locally similar queries. We thus propose a Query Contrast mechanism to explicitly enhance queries towards their best-matched GTs over all unmatched query predictions. This is achieved by the construction of positive and negative GT-query pairs for each GT, and a contrastive loss to enhance positive GT-query pairs against negative ones based on feature similarities. ConQueR closes the gap of sparse and dense 3D detectors, and reduces up to \textbf{$\sim$60\%} false positives. Our single-frame ConQueR achieves new state-of-the-art (sota) 71.6 mAPH/L2 on the challenging Waymo Open Dataset validation set, outperforming previous sota methods (\emph{e.g.}, PV-RCNN++) by over \textbf{2.0} mAPH/L2.

 \end{abstract}

\section{Introduction}
\label{sec:intro}

3D object detection from point clouds has received much attention in recent years~\cite{zhou2018voxelnet,shi2019pointrcnn,shi2020pv,yin2021center,fan2022fully} as its wide applications in autonomous driving, robots navigation, etc. 
State-of-the-art 3D detectors~\cite{shi2021pv,shi2022pillarnet,fan2022fully,zhou2022centerformer} still adopt dense predictions with post-processing (\emph{e.g.}, NMS~\cite{canny1986computational}) to obtain final sparse detections. This \emph{indirect} pipeline usually involves many hand-crafted components (e.g., anchors, center masks) based on human experience, which involves much effort for tuning, and prevents dense detectors from being optimized end-to-end to achieve optimal performance.
Recently, DETR-based 2D detectors~\cite{carion2020end,zhu2020deformable,sun2021sparse,zhang2022dino} show that transformers with \emph{direct} sparse predictions can greatly simplify the detection pipeline, and lead to better performance. 
However, although many efforts~\cite{Misra_2021_ICCV,nguyen2022boxer,bai2022transfusion} have been made towards \emph{direct} sparse predictions for 3D object detection, because of the different characteristics of images and point clouds (\emph{i.e.}, dense and ordered images \emph{v.s.} sparse and irregular points clouds), 
performance of sparse 3D object detectors still largely lags behind state-of-the-art dense detectors.

\begin{figure}[t]
  \centering
   \includegraphics[width=0.95\linewidth]{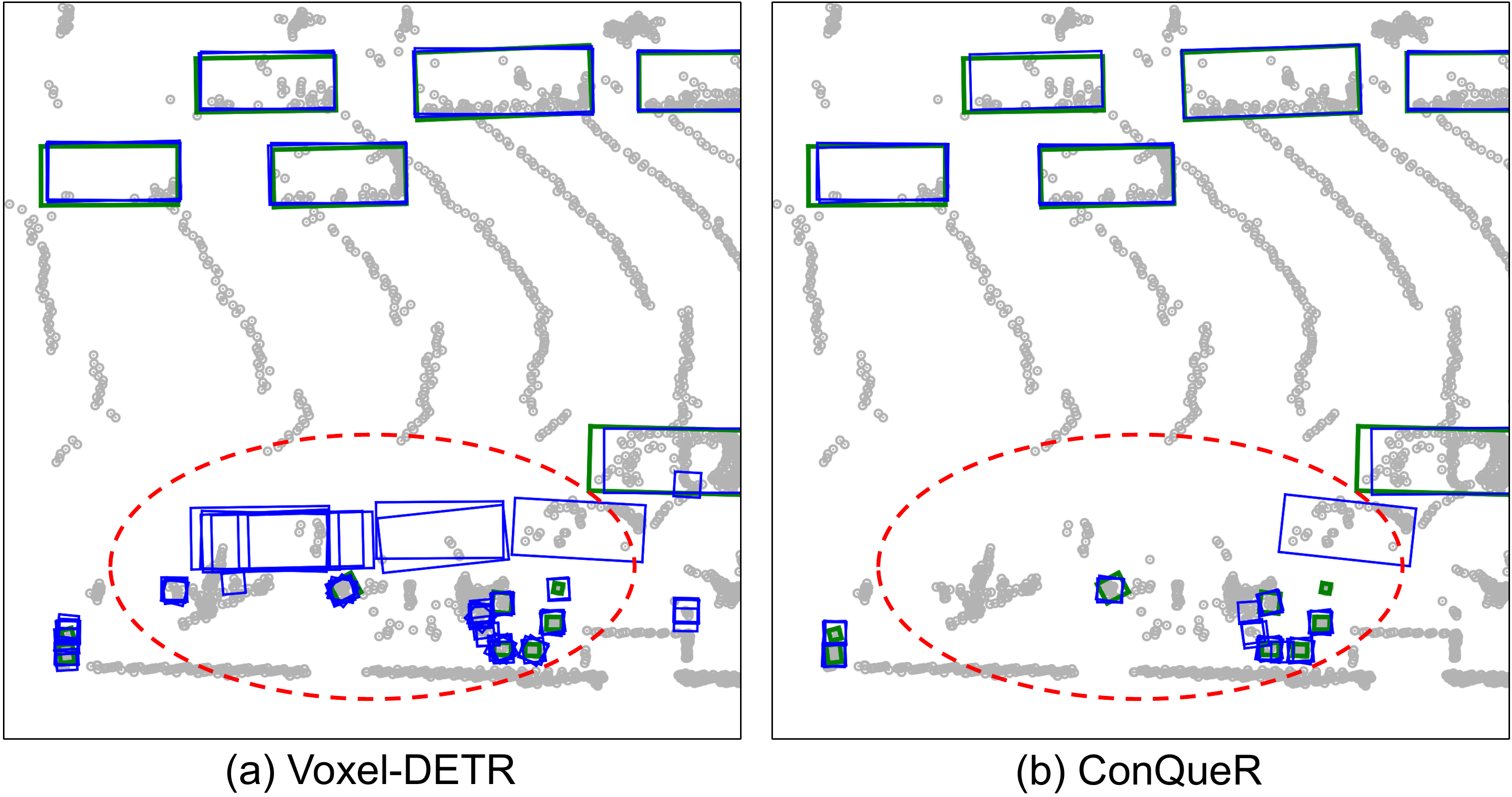}
   \caption{Comparison of our baseline Voxel-DETR and ConQueR. GTs ({\color{green}{green}}) and predictions ({\color{blue}{blue}}) of an example scene in the WOD is visualized. Sparse predictions of Voxel-DETR still contain many highly overlapped false positives (in the red dashed circle), while ConQueR can generate much sparser predictions.}
   \label{fig:motivation}
   \vspace{-3mm}
\end{figure}

To achieve direct sparse predictions, DETRs usually adopt a set of object queries~\cite{carion2020end,zhu2020deformable,sun2021sparse,zhang2022dino,nguyen2022boxer,bai2022transfusion}, and resort to the one-to-one Hungarian Matching~\cite{kuhn1955hungarian} to assign ground-truths (GTs) to object queries. 
However, to guarantee a high recall rate, those detectors need to impose much more queries than the actual number of objects in a scene. For example, recent works~\cite{nguyen2022boxer,bai2022transfusion} select top-$300$ scored query predictions to cover only $\sim$40 objects in each scene of Waymo Open Dataset (WOD)~\cite{sun2020scalability}, while 2D DETR detectors~\cite{carion2020end,zhu2020deformable,sun2021sparse,zhang2022dino} use 10$\times$ more predictions than the average GT number of MS COCO~\cite{lin2014microsoft}. As shown in Fig.~\ref{fig:motivation}(a), we visualize an example scene by a baseline DETR-based 3D detector, named Voxel-DETR, which shows its top-$300$ scored predictions.  
Objects are generally small and densely populated in autonomous driving scenes,  while 3D DETRs adopt the same fixed top-$N$ scored predictions as 2D DETRs, and lack a mechanism to handle such small and dense objects. Consequently, they tend to generate densely overlapped false positives (in the red-dashed circle), harming both the accuracy and \emph{sparsity}~\cite{sun2021sparse,roh2021sparse} of final predictions. 

We argue the key reason is that the Hungarian Matching in existing 3D DETRs only assigns each GT to its best matched query, while all other unmatched queries near this GT are not effectively suppressed. For each GT, the one-to-one matching loss solely forces all unmatched queries to predict the same ``no-object'' label, and the best matched query are supervised \emph{without} considering its relative ranking to its surrounding unmatched queries. This design causes the detectors to be insufficiently supervised in discriminating similar query predictions for each GT, leading to duplicated false positives for scenes with densely populated objects. 

To overcome the limitations of current supervision,  we introduce a simple yet novel Query Contrast strategy to explicitly suppress predictions of all unmatched queries for each GT, and simultaneously enhance the best matched query to generate more accurate predictions in a contrastive manner.
The Query Contrast strategy is integrated into our baseline Voxel-DETR, which consists of a sparse 3D convolution backbone to extract features from voxel grids, and a transformer encoder-decoder architecture with a bipartite matching loss to directly generate sparse predictions.
Our Query Contrast mechanism involves the construction of positive and negative GT-query pairs, and the contrastive learning on all GT-query pairs to supervise both matched and unmatched queries with knowledge of the states of their surrounding queries. Such GT-query pairs are directly created by reusing the Hungarian Matching results: each GT and its best matched query form the positive pair, and all other unmatched queries of the same GT then form negative pairs.
To quantitively measure the similarities of the GT-query pairs, we formulate the object queries to be the same as GT boxes (\emph{i.e.}, using only box categories, locations, sizes and orientations), such that GTs and object queries can be processed by the same transformer decoder, and embedded into a unified feature space to properly calculate their similarities.
Given the GT-query similarities, we adopt the contrastive learning loss~\cite{chen2020simple,he2020momentum,zhu2020eqco} to effectively enhance the positive (matched) query's prediction for each GT, and suppress those of all its negative queries at the same time. Moreover, to further improve the contrastive supervision, we construct multiple positive GT-query pairs for each GT by adding small random noises to the original GTs, which greatly boost the training efficiency and effectiveness. 

The resulting sparse 3D detector, named \textbf{Que}ry \textbf{Con}trast Voxel-DET\textbf{R} (\textbf{ConQueR}), significantly improves the detection performance and sparsity of final predictions, as shown in Fig.~\ref{fig:motivation}(b). Moreover, ConQueR abandons the fixed top-$N$ prediction scheme and enables to output a vary number of predictions for different scenes. ConQueR reduces up to {\bf $\sim$60\%} false positives and sets new records on the challenging Waymo Open Dataset (WOD)~\cite{sun2020scalability}. Contributions are summarized as bellow:
\begin{enumerate}
	\item We introduce a novel Query Contrast strategy into DETR-based 3D detectors to effectively eliminate densely overlapped false positives and achieve more accurate predictions. 
	\item We propose to construct multi-positive contrastive training, which greatly improve the effectiveness and efficiency of our Query Contrast mechanism.
	\item Our proposed sparse 3D detector ConQueR closes the gap between sparse and dense 3D detectors, and sets new records on the challenging WOD benchmark.
\end{enumerate}

\section{Related Works}

\paragraph{End-to-End 2D Object Detection.} End-to-end object detection aims to generate final sparse predictions without non-differentiable components like NMS. RelationNet~\cite{hu2018relation} proposes an object relation module and DETR~\cite{carion2020end} greatly simplifies the detection pipeline by removing many hand-crafted components like anchors, NMS, etc. DETR introduce a set of object queries and resorts to the Hungarian Matching to associate each GT with the query predictions of minimal matching cost, and selects top-$N$ scored predictions for inference. \cite{wang2021end,sun2021sparse} also reveal that one-to-one matching is the key to achieve sparse predictions. Following works~\cite{zhu2020deformable,meng2021conditional,wang2022anchor,li2022dn,li2022dn,jia2022detrs} improves DETR in many aspects including query design, convergence speed, and performance, surpassing CNN-based dense detectors~\cite{zhang2020bridging,zhu2020autoassign,ge2021ota} by a large margin. However, they still need to select a fixed number of predictions as final results, no matter how many objects are there in an image.  
Recently, DINO-DETR~\cite{zhang2022dino} introduces a ``contrastive'' denoising training strategy. It creates positive and negative GTs conceptually, and supervise these GTs with different targets separately, which has no relation with contrastive learning.

\begin{figure*}[t]
  \centering
   \includegraphics[width=0.9\linewidth]{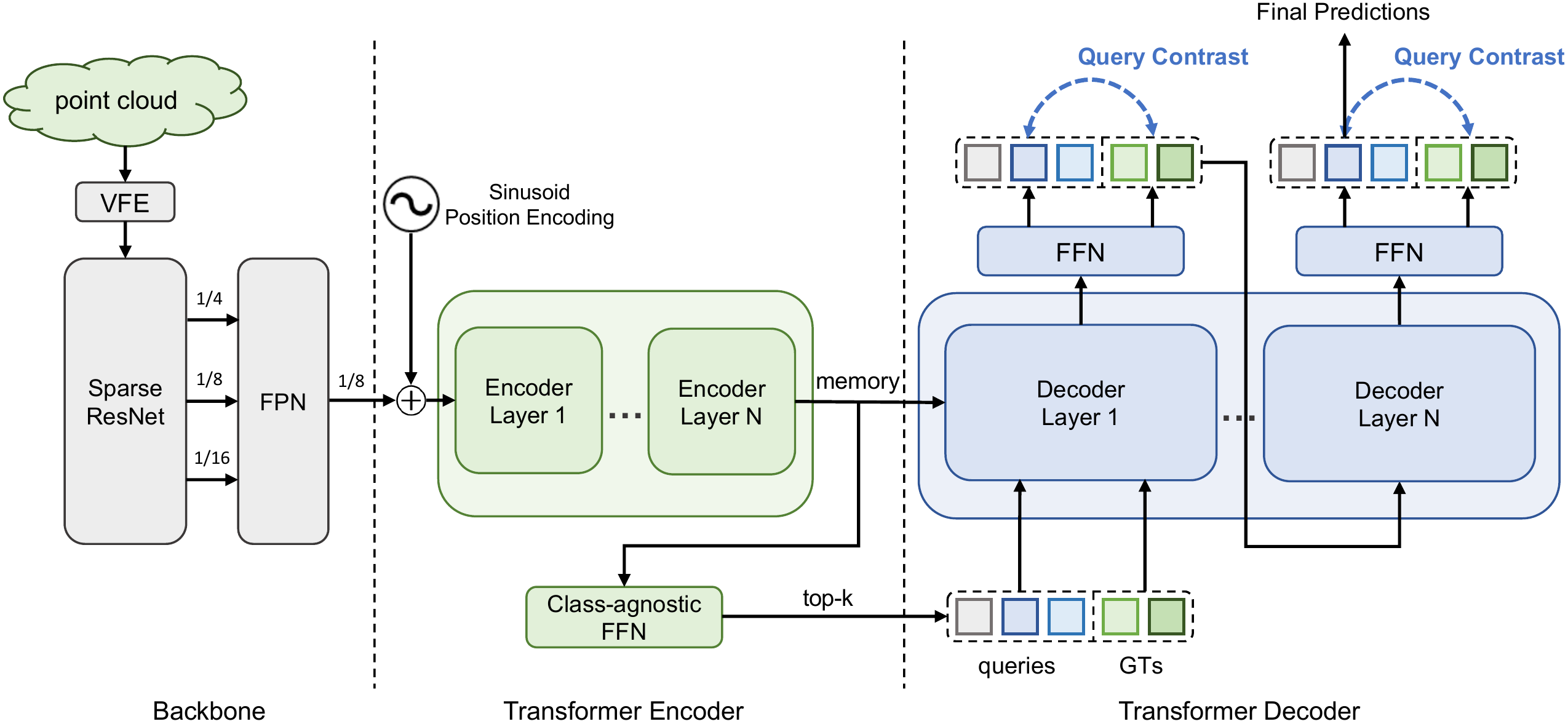}
   \caption{Overall pipeline of the proposed ConQueR. It consists of a 3D Sparse ResNet-FPN backbone to extract dense BEV features, and a transformer encoder-decoder architecture with one-to-one matching. Top-$k$ scored object proposals from a class-agnostic FFN form the object queries to input to the transformer decoder. During training, GTs (noised) are concatenated with object queries to input to the transformer decoder to obtain unified embeddings, which are then used for Query Contrast at each decoder layer. During inference, Top-scored predictions from the last decoder layer are kept as final sparse predictions. ``VFE'' denotes the voxel feature extractor in ~\cite{yan2018second,zhu2019class,yin2021center}.
   }
   \label{fig:contrastive_voxel_detr}
\end{figure*}

\vspace{-2mm}
\paragraph{3D Object Detection from Point Clouds.} State-of-the-art 3D detectors usually adopts voxel-based~\cite{shi2020pv,yin2021center,shi2021pv,shi2022pillarnet}, range-view~\cite{sun2021rsn,tian2022fully} or point-based~\cite{fan2022fully,yang20203dssd} paradigms to convert raw point clouds into dense feature representations, followed by detection heads to generate dense predictions and resort to NMS to filter out low-quality predictions. Many attempts have also been made to incorporate transformer architectures~\cite{sheng2021improving,mao2021voxel,zhou2022centerformer,Sun2022SWFormerSW} into 3D object detection, but they still rely on post-processing. Others~\cite{nguyen2022boxer,bai2022transfusion} make a step further to use the one-to-one matching loss to achieve direct sparse 3D predictions. \cite{nguyen2022boxer} proposes Box-Attention, a variant of deformable attention to better capture local informations and applies it to 3D object detection. \cite{bai2022transfusion} introduce image features into a decoder-only architecture to enhance query features. However, their performance still largely lags behind state-of-the-art dense 3D detectors.

\vspace{-2mm}
\paragraph{Contrastive Learning for Object Detection.} Contrastive learning aims to learn an embedding space such that similar data pairs stay close while dissimilar ones are far apart. ~\cite{hadsell2006dimensionality} proposes to learn representations by contrasting positive pairs against negative ones. The popular InfoNCE loss~\cite{oord2018representation} uses categorical cross-entropy loss to learn such an embedding space. Following works~\cite{chen2020simple,he2020momentum,caron2020unsupervised} demonstrate the superiority of contrastive learning on providing pre-trained weights for downstream tasks (\emph{e.g.}, 2D detection). Few works explore the use of contrastive loss in object detection. ~\cite{lang2021contrastive} introduces semantically structured embeddings from knowledge graphs to alleviate misclassifications. ~\cite{yao2021g} conducts contrastive distillation between different feature regions to better capture teacher's information. As far as we know, we are the first to introduce the contrastive learning process into DETR-based detectors.

\section{Query Contrast Voxel-DETR (ConQueR)}
\label{sec:method}

State-of-the-art 3D detectors usually generate dense object predictions, which require many hand-designed components (\emph{e.g.}, anchors, box masks) based on prior knowledge, and resort to post-processing to filter out low-quality and duplicated boxes. This \emph{indirect} pipeline hinders the detectors from being optimized end-to-end and achieving optimal performance. 3D DETRs aim at streamlining these hand-crafted modules, and \emph{directly} generating sparse predictions via the transformer architecture and one-to-one matching loss, but they still cannot compete with state-of-the-art dense 3D detectors and face the problem of highly overlapped false positives, as shown in Fig.~\ref{fig:motivation}(a). To solve these challenges, we first introduce our competitive DETR-based 3D framework, named Voxel-DETR in Sec.~\ref{sec:voxel_detr}, and present the Query Contrast strategy to tackle with the duplicated false positives and further improve the detection performance in Sec.~\ref{sec:cns}. 

\subsection{Voxel-DETR}
\label{sec:voxel_detr}
 
As illustrated in Fig.~\ref{fig:contrastive_voxel_detr}, Voxel-DETR consists of a 3D backbone, an encoder-decoder transformer architecture, and a set-matching loss to achieve direct sparse predictions.

\noindent {\bf Backbone.} Point cloud is rasterized into sparse voxel grids and fed into a 3D Sparse ResNet~\cite{he2016deep} backbone network to extract sparse 3D features. These features are transformed into dense Bird Eye View (BEV) feature maps, followed by an FPN~\cite{lin2017feature} to extract multi-scale features.

\noindent {\bf Transformer.} 
The  encoder-decoder transformer is similar to the two-stage Deformable-DETR~\cite{zhu2020deformable}. The $8\times$ downscaled BEV features from the FPN are input to the transformer encoder, which consists of 3 encoder layers. 
Considering the characteristics of 3D detection from point clouds (\emph{i.e.}, all objects are relatively small and densely distributed),
we adopt BoxAttention~\cite{nguyen2022boxer}, which applies spatial in-box constraints to Deformable Attention~\cite{zhu2020deformable}, to perform local self-attention.
A class-agnostic feed-forward network (FFN) head is used to generate initial object proposals from encoder features. Top-$k$ scored box proposals are selected as object queries to input to the 3-layer transformer decoder. Decoder layers conduct inter-query self-attention and cross-attention between query and encoder features, followed by prediction heads to perform iterative box refinement~\cite{zhu2020deformable}. Predicted query boxes from the previous decoder layer's FFN head are transformed by a 3-layer MLP and added with the updated query features (initialized as zero) from the previous decoder layer.

\noindent {\bf Losses.} During training, all FFN prediction heads use the Hungarian Matching to assign GTs to object queries. The detection loss $\mathcal{L}_{det}$ consists of a focal loss~\cite{lin2017focal} for classification, a smooth L1 loss and a 3D GIoU loss for box regression:
\begin{equation}
\label{eq:det_loss}
	\mathcal{L}_{\rm det} = \alpha \mathcal{L}_{\rm focal} + \beta \mathcal{L}_{\rm l1} + \gamma \mathcal{L}_{\rm GIoU},
\end{equation}
where $\alpha, \beta, \gamma$ are hyper-parameters to balance the loss terms. 
During inference, top-$N$ scored predictions from the last decoder layer are kept as the final sparse detections.

\subsection{Query Contrast}
\label{sec:cns}

\begin{figure}[t]
  \centering
   \includegraphics[width=1.0\linewidth]{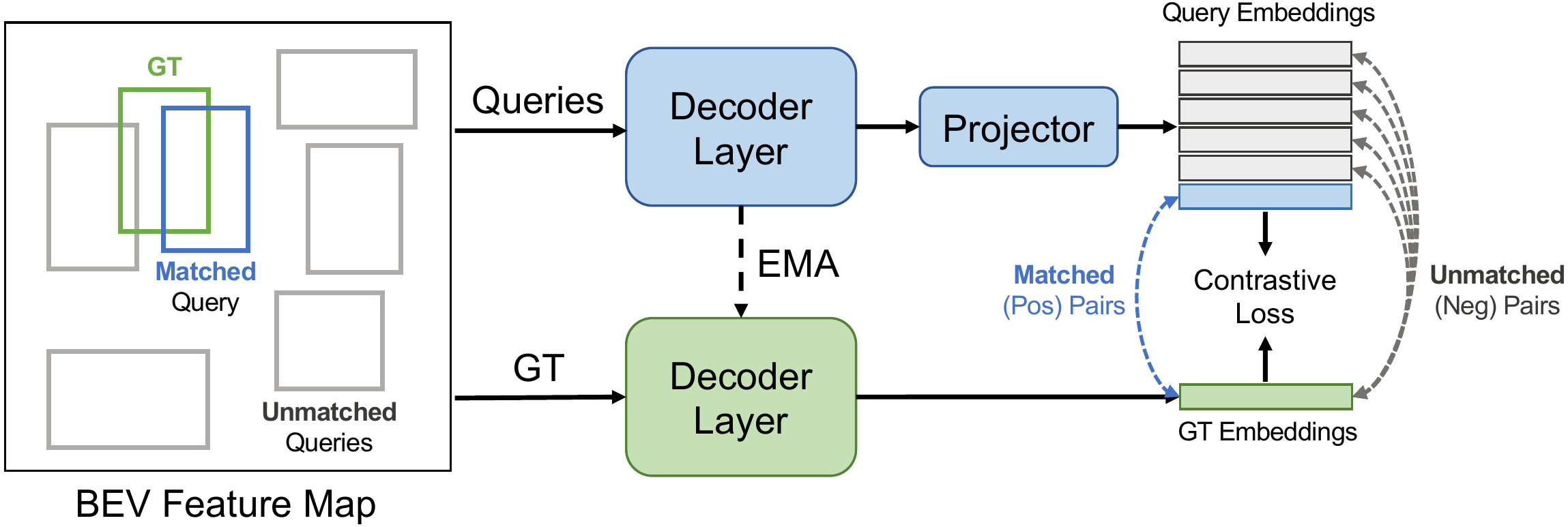}
   \caption{Illustration of Query Contrast. Given the GT ({\color{green}green}), Hungarian Matching gives its best matched ({\color{blue}blue}) and all other unmatched ({\color{gray}gray}) object queries. Query embeddings are projected by an extra MLP to align with GT embeddings. The contrastive loss is applied to all positive and negative GT-query pairs based on their feature similarities. 
   }
   \label{fig:contrastive_loss}
 \vspace{-3mm}
\end{figure}
 
Although Voxel-DETR already achieves satisfactory performance, its top-$N$ scored predictions still suffer from densely overlapped false positives (as shown in Fig.~\ref{fig:motivation}(a)). To tackle this problem, 
we present a novel Query Contrast mechanism (depicted in Fig.~\ref{fig:contrastive_loss}) to explicitly enhance each GT's best matched query over unmatched ones. We first construct positive and negative GT-query pairs for each GT, which are then processed by each decoder layer to generate aligned GT and query embeddings. To promote the positive queries' similarity towards a GT against negative ones, the contrastive loss is applied at each decoder layer. 

\vspace{-2mm}
\paragraph{Construction of positive/negative GT-query pairs.}
To determine queries to be enhanced or suppressed for each GT, we first construct positive and negative GT-query pairs by reusing the Hungarian Matching results (used for Eq.(\ref{eq:det_loss})), which is naturally compatible with our Voxel-DETR framework. Given a GT, the query with the minimal matching cost forms a positive pair with the GT, all other queries and this GT then form negative GT-query pairs. These GT-query pairs help to identify the object queries that need to be further enhanced or suppressed in our Voxel-DETR. Motivated by the SwAV~\cite{caron2020unsupervised} that incorporates multiple image crops to form multiple positive pairs to boost the training process, we further add small noises of different magnitudes on each GT to generate multiple noised GT copies. The multiple noised GT copies then form additional GT-query pairs with the same positive/negative query partitions as original GTs.

In practice, if a noised copy deviates too much from its original GT, the noised GT-query pairs would harm the contrastive training process. However, finding proper noise magnitudes is rather laboursome and cannot generalise well across scenarios. We thus add an auxiliary GT de-noising loss similar to that in DN-DETR~\cite{li2022dn} to obligate the detector to recover the original GT from its noised versions, which ensures that the noised GT copies would not diverge. Note that the ``noising-denoising'' step alone only has marginal effects to detection performance, while our multi-positive Query Contrast based on the noised GT copies leads to superior detection performance, as shown in our ablation studies.

\vspace{-4mm}
\paragraph{Contrast positive pairs against negative pairs.}
Before applying supervisions to the positive and negative GT-query pairs, we need to quantitatively measure the similarities of these pairs.
However, simple geometric metrics (\emph{e.g.}, IoU) cannot sufficiently model the similarities between GTs and queries (\emph{i.e.}, category, appearance, location, size, etc.). We thus propose to embed GTs and queries into a latent space for comprehensive similarity measurement. In our Voxel-DETR, object queries are formulated as proposal boxes (\emph{i.e.} object category, box location, size, and orientation). Therefore, the transformer decoder can naturally be used to encode both GTs and queries into feature embeddings at a chosen layer. We simply select the output layer of the FFN prediction head after each decoder layer (as shown in Fig.~\ref{fig:contrastive_voxel_detr}), followed by a shared MLP for similarity estimation.

However, we observe that the distributions of GT objects and query boxes can be quite different: GTs have no overlap with each other and generally distribute following the roadmap layouts, while queries might correspond to densely overlapped boxes and show up at random locations. As the transformer decoder utilizes self-attention to capture inter-box relations, the different distributions of GTs and query boxes would greatly affect estimation of their similarities. To mitigate the distribution gap, we adopt an extra MLP to project query features to align with GTs' latent space (the ``Projector'' in Fig.~\ref{fig:contrastive_loss}). With the aligned GT and query embeddings, we estimate all positive and negative GT-query pairs' similarities with cosine similarity metric, and adopt the InfoNCE loss~\cite{oord2018representation} to encourage the best matched query to generate more accurate predictions towards its assigned GT, and force all other unmatched queries to deviate away. Moreover, to obtain more stable GT representations for supervising queries, we adopt an exponential moving average (EMA) copy for each decoder layer to embed GTs, which is shown to be effective in our ablations.

Assume that for the $i$-th GT in a point cloud scene, we add $T$ different noises and denote the noised GT embeddings as $\{b^1_i, b^2_i, ... , b^T_i\}$, and denote $K$ query embeddings as $\{q_1, q_2, ... , q_K\}$. Suppose that the Hungarian Matching assigns the $i$-th GT to the $j$-th query, then our Query Contrast loss for the $i$-th GT  $\mathcal{L}^{\rm QC}_i$ can be formulated as 
\begin{equation}
	\mathcal{L}^{\rm QC}_i = -\sum_{t=1}^{T} \log \left( \frac{\exp( \cos(b^t_i, g(q_j) ) / \tau  )}{\sum_{k=1}^{K}\exp(\cos(b^t_i, g(q_k)) / \tau)} \right),
\label{eq:qc}
\end{equation}
where $\tau$ is the temperature coefficient, and $g(\cdot)$ denotes the extra MLP projector to align query features to GTs'.
As shown in Fig.~\ref{fig:contrastive_voxel_detr}, the Query Contrast loss is adopted at every decoder layer. 

During inference, we abandon the widely adopted top-$N$ scored prediction strategy and use a score threshold (\emph{e.g.}, 0.1) to filter out low-quality query predictions. Query Contrast works quite well on suppressing similar query predictions in local neighborhoods, as shown in Fig.~\ref{fig:motivation}(b). ConQueR greatly boosts the detection accuracy, and reduces up to \textbf{$\sim$60\%} false positives.

\vspace{-1mm}
\paragraph{Discussion: Why does Query Contrast improve DETR-based 3D detectors?}

As discussed in Sec.~\ref{sec:intro}, current detection losses (\emph{i.e.}, focal loss for classification, smooth L1 and GIoU loss for regression) supervise each query without considering its surrounding queries, which lack supervision to train detectors to discriminate similar object queries especially in local regions. The proposed Query Contrast strategy tackles this issue by constructing a contrastive objective to supervise \emph{all} queries simultaneously. As suggested in Eq.(\ref{eq:qc}), for each GT object, the detector is instructed to identify the best matched query, and is forced to learn to differentiate it from all other unmatched counterparts, even if some of them are highly overlapping with the best matched query. As a result, all unmatched queries are trained to deviate from the GT, thus the duplicated false positives in our baseline Voxel-DETR can be effectively suppressed.

Another core design of our Query Contrast is to encode the GTs and queries into a unified learnable latent space. GT objects are encoded to provide better forms of supervision for both matched and unmatched queries. Previous works~\cite{hao2020labelenc,zhang2022lgd} in 2D object detection also show that encoding labels into feature embeddings to serve as extra supervision can perform better than the common hand-designed learning targets (\emph{i.e.}, classification logits and regression offsets), but they generally work in a knowledge distillation (KD) manner, which cannot be utilized to supervise negative queries. In contrast, our contrastive loss does not force matched queries to approach GTs directly, but encourages them to be ``closer'' to their corresponding GT embeddings than other close-by duplicated queries. Note that in our Query Contrast mechanism, GT embeddings are processed in an off-line manner and encoded into a unified space as queries', which serve as a type of supervision and force the detector to generate more similar query features as GTs'. 

According to our experiments, the proposed Query Contrast strategy can not only suppress those duplicated false positives, but also contribute to better detection performance, which are consistent with the above discussions.

\section{Experiments}

ConQueR is mainly evaluated on the Waymo Open Dataset~\cite{sun2020scalability} (WOD) benchmark using the official detection metrics:  mAP and mAPH (mAP weighted by heading) for Vehicle (Veh.), Pedestrian (Ped.), and Cyclist (Cyc.). The metrics are further splitted into two difficulty levels according to the point numbers in GT boxes: LEVEL\_1 ($>$5) and LEVEL\_2 ($\geq$1). We conduct ablation studies on the \textit{validation} set, and compare with state-of-the-art detectors on both \textit{validation} and \textit{test} set.

\begin{table*}[t]
\footnotesize
\centering
\begin{tabular}{l|>{\columncolor[gray]{0.95}}c|c|c|c|c|c|c}
  \specialrule{1pt}{0pt}{1pt}
\toprule
\multicolumn{1}{c|}{\multirow{2}{*}{Methods}} & \multirow{2}{*}{\shortstack[1]{mAP/mAPH\\ L2}} & \multicolumn{2}{c|}{\emph{Vehicle} 3D AP/APH} & \multicolumn{2}{c|}{\emph{Pedestrian} 3D AP/APH} & \multicolumn{2}{c}{\emph{Cyclist} 3D AP/APH}\\
 & L2 &   L2      &   L1     &   L2      &   L1  &   L2     &   L1\\
\midrule
\multicolumn{8}{l}{\textbf{Dense Detectors}}\\
CenterPoint$_{\rm ts}$~\cite{yin2021center} & -/67.4 & -/67.9 & -/- & -/65.6 & -/- & -/68.6-/- & -/- \\
PV-RCNN~\cite{shi2020pv}   &66.8/63.3& 69.0/68.4 & 77.5/76.9 & 66.0/57.6  &  75.0/65.6 & 65.4/64.0 & 67.8/66.4 \\
AFDetV2~\cite{hu2022afdetv2} & 71.0/68.8 & 69.7/69.2 & 77.6/77.1 & 72.2/67.0 & 80.2/74.6 & 71.0/70.1 & 73.7/72.7\\
SST\_TS~\cite{fan2022embracing} & -/- & 68.0/67.6  & 76.2/75.8 & 72.8/65.9 & 81.4/74.1 & -/- & -/- \\
SWFormer~\cite{Sun2022SWFormerSW} &  -/- & 69.2/68.8 & 77.8/77.3 & 72.5/64.9 & 80.9/72.7 & -/- & -/- \\
PillarNet-34~\cite{shi2022pillarnet} & 71.0/68.5 & \underline{70.9}/\textbf{70.5} & \underline{79.1}/\underline{78.6} & 72.3/66.2 & 80.6/74.0 & 69.7/68.7 & 72.3/71.2 \\
CenterFormer\cite{zhou2022centerformer} & 71.2/69.0 & 70.2/69.7 &  75.2/74.7 & 73.6/\underline{68.3} & 78.6/73.0 & 69.8/68.8 & 72.3/71.3 \\
PV-RCNN++~\cite{shi2021pv}  & 71.7/69.5 & 70.6/\underline{70.2} & \textbf{79.3/78.8} &  73.2/68.0  &  \underline{81.3}/\underline{76.3} & 71.2/70.2 & 73.7/72.7\\
\midrule
\multicolumn{8}{l}{\textbf{Sparse Detectors}}\\
BoxeR-3D & -/- & 63.9/63.7 & 70.4/70.0 & 61.5/53.7 & 64.7/53.5 & -/-  & 50.2/48.9 \\
TransFusion-L & -/64.9 & -/65.1 & -/- & -/63.7 & -/- & -/65.9 & -/- \\
Voxel-DETR (ours) & 68.8/66.1 & 67.8/67.2 & 75.4/74.9 & 69.7/63.1 & 77.6/70.5 & 69.0/67.9 & 71.7/70.5\\ 
ConQueR (ours) & 70.3/67.7 & 68.7/68.2 & 76.1/75.6 & 70.9/64.7 & 79.0/72.3 & 71.4/70.1 & 73.9/72.5 \\
ConQueR \dag (ours) & \underline{73.1}/\underline{70.6} & \textbf{71.0}/\textbf{70.5} &  78.4/77.9  &  \underline{73.7}/68.1  &  80.9/75.2 &  \underline{74.5}/\underline{73.3} & \underline{77.3}/\underline{76.1} \\
ConQueR \ddag (ours) & \textbf{74.0}/\textbf{71.6} & \textbf{71.0}/\textbf{70.5} & 78.4/77.9 &  \textbf{75.8}/\textbf{70.1} &  \textbf{82.4}/\textbf{76.6}  &  \textbf{75.2}/\textbf{74.1} & \textbf{77.5}/\textbf{76.4} \\
\bottomrule
  \specialrule{1pt}{1pt}{0pt}
\end{tabular}
\vspace{-1mm}
\caption{Performances on the WOD \emph{validation} split. All models take single-frame input with the same range, no pre-training or ensembling is required. $\dag$ denotes using the 2$\times$ wider ResNet~\cite{zagoruyko2016wide} with $1/4$ downscaled BEV feature map in our backbone. $\ddag$ denotes conducting NMS on pedestrians and cyclists. \textbf{Bold} denotes the best entries, and \underline{underline} denotes the second-best entries. $_{\rm ts}$ denotes the two-stage model.
}
 
\label{tab:sota}
\end{table*}

\subsection{Implementation Details}

\noindent {\bf Training.} We follow common practice as previous voxel-based methods~\cite{shi2020pv,yin2021center,shi2021pv,shi2022pillarnet} to use point cloud range of $[-75.2m, 75.2m]\times[-75.2m, 75.2m]\times[-2.0m, 4.0m]$ with voxel size $[0.1m, 0.1m, 0.15m]$ in $x$, $y$, and $z$-axes respectively. The same set of augmentations (\emph{i.e.}, GT-Aug, flip, rotation, scaling) are adopted following the previous works~\cite{yin2021center}. We follow ~\cite{wang2021pointaugmenting,bai2022transfusion} to use the ``fade-strategy'' to drop GT-Aug at the last epoch to avoid overfitting. Both our baseline Voxel-DETR and ConQueR are trained for 6 epochs unless otherwise specified. We use the OneCycle~\cite{smith2019super} learning rate scheduler and AdamW~\cite{loshchilov2017decoupled} optimizer with maximal learning rate 0.001. 

\noindent{\bf Network.} For the 3D backbone in Fig.~\ref{fig:contrastive_voxel_detr}, we use the same architecture as ResNet-18~\cite{he2016deep} but use sparse 3D convolutions~\cite{graham2017submanifold} to replace the 2D ones. No pre-trained weights are used. The same FPN structure as RetinaNet~\cite{lin2017focal} is used to obtain multi-scale BEV features. For simplicity, we only use the 8$\times$ downscaled features as input to the transformer, which adopts 3 encoder layers and 3 decoder layers for computation efficiency. We select top-$1000$ scored query predictions from the encoder's class-agnostic prediction head as object queries. We adopt top-N (\emph{e.g.}, 300) scored predictions, or score threshold (\emph{e.g.}, $\geq$ 0.1) during inference. We set $\alpha=1$, $\beta=4$, $\gamma=2$ in Eq.~(\ref{eq:det_loss}). For the proposed Query Contrast, we use $\tau=0.7$ in Eq.~(\ref{eq:qc}), and adopt $T=3$ noising groups with a maximal box noise ratio of 0.4~\cite{li2022dn}, and label noise ratio of 0.5~\cite{li2022dn}. Category labels are simply encoded as one-hot embeddings rather than the learnable embeddings in DN-DETR~\cite{li2022dn}.

\subsection{Main Results}

For fair comparison, all methods included use the same point cloud input range, do not use any pre-trained weights, test-time augmentation or model ensembling.

\vspace{-4mm}
\paragraph{Performance.}

As shown in Table~\ref{tab:sota}, state-of-the-art 3D detectors are divided into dense and sparse categories according to whether they can \emph{directly} generate sparse detections. Our sparse detector ConQueR sets new records on \emph{all categories} of the WOD \emph{validation} set. ConQueR with direct sparse predictions (the second-last entry) achieves $\sim$1.0 mAPH/L2 higher than the previous best single-frame model PV-RCNN++~\cite{shi2021pv}, and is over 3.0 mAPH/L2 higher than the popular anchor-free CenterPoint~\cite{yin2021center}. Notably, ConQueR demonstrates overwhelming performance on pedestrians and cyclists, outperforming previous best methods by $\sim$\textbf{2.0} APH/L2, which shows the effectiveness of our Query Contrast strategy especially for densely populated categories. The significant performance improvements can also be validated on the WOD \emph{test} set in Table~\ref{tab:sota_test}. Moreover, ConQueR surpasses previous best sparse detectors TransFusion-L by \textbf{$\sim$6.0} mAPH/L2, closing the performance gap between sparse and dense 3D detectors. When compared with our baseline Voxel-DETR, the proposed Query Contrast mechanism brings over \textbf{1.6} mAPH/L2 without any extra inference cost. Besides, our baseline Voxel-DETR with only 6 epochs of training outperforms previous sparse 3D detectors, and achieves comparable performance with CenterPoint (36-epoch training) with only $1/6$ GPU hours. In addition, ConQueR has an inference latency of 70ms (46ms for CenterPoint)\footnote{Latency is measured with batch size 1 on NVIDIA A100 GPU.}.

Although ConQueR with \emph{direct} sparse predictions already achieves state-of-the-art performance, we find that applying NMS onto ConQueR's sparse predictions can further improve small and densely populated categories such as pedestrians, while NMS causes $\sim$1.2 APH/L2 performance drop on the well-trained vehicles (as shown in Appendix.~\ref{appdenx:a}). This is also the case with our baseline Voxel-DETR. We speculate this is caused by the learning difficulties inherent in the data for extremely similar queries (as shown in Fig.~\ref{fig:motivation}(b)) . We thus report ConQueR's performance after conducting NMS on pedestrians and cyclists (the last entry of Table~\ref{tab:sota}).\footnote{FSD~\cite{fan2022fully} adopts larger point cloud ranges and requires a point segmentation pre-train weights. It is $\sim$1.0 mAPH/L2 lower than our ConQueR.}

\begin{table}[t]
\footnotesize
\centering
\begin{tabular}{l|>{\columncolor[gray]{0.95}}c|c|c|c}
\hline
\multicolumn{1}{c|}{Methods}  & All  & Veh. & Ped. & Cyc. \\ \hline
CenterPoint~\cite{yin2021center}  & 69.0 & 71.9 & 67.0 & 68.2 \\ \hline
PV-RCNN++~\cite{shi2021pv}    & 70.2 & 73.5 & 69.0 & 68.2 \\ \hline
AFDetv2~\cite{hu2022afdetv2}      & 70.0 & 72.6 & 68.6 & 68.7 \\ \hline
PillarNet-34~\cite{shi2022pillarnet} & 69.6 & \bf74.7 & 68.5 & 65.5 \\ \hline
ConQueR (Ours)      & \bf72.0 & 73.3 & \bf70.9 & \bf71.9 \\ \hline
\end{tabular}
\caption{Single-frame performance comparisons on the WOD \emph{test} set. APH/L2 results are reported.}
\label{tab:sota_test}
\vspace{-2mm}
\end{table}

\begin{table}[t]
\footnotesize
\centering
\begin{tabular}{l|c|c|c|c}
\hline
\multicolumn{1}{c|}{Methods}   & Preds/Scene & Veh.  & Ped.  & Cyc.                         \\ \hline
$\text{CenterPoint}_{\rm nms}$    & 192         & 66.4 & 62.9 & 67.9              \\ \hline
$\text{Transfusion}_{\rm topN}$    & 300         & 65.1 & 63.7 & 65.9              \\ \hline \hline
$\text{Voxel-DETR}_{\rm topN}$     & 300         & 67.1 & 63.0 & 67.8              \\ \hline
$\text{Voxel-DETR}_{\rm score}$   & 222         & 67.2  &  63.1  & 67.9            \\ \hline \hline
$\text{ConQueR}_{\rm topN}$   & 300         & 68.0 & 64.6 & 70.0            \\ \hline
$\text{ConQueR}_{\rm score}$ & 131         & 68.2 & 64.7 & 70.1       \\ \hline
$\text{ConQueR}_{\rm score}$ \dag & 122         & \bf70.5 & \bf68.1 & \bf73.3       \\ \hline
\end{tabular}
\caption{Sparsity of final predictions. APH/L2 results are reported on the WOD \textit{validation} set. The subscripts of each entry denotes the way they obtain final predictions. For example, CenterPoint$_{\rm nms}$ uses NMS to filter out duplicated boxes, and Voxel-DETR$_{\rm topN}$ denotes it uses top-$N$ scored proposals as final predictions, while ConQueR$_{\rm score}$ denotes that using score thresholding to generate final sparse predictions. $\dag$ denotes our best model in Table~\ref{tab:sota}.}
\label{tab:sparsity}
\vspace{-4mm}
\end{table}

\vspace{-2mm}
\paragraph{Sparsity.} 
Apart from the performance improvements on the WOD official metrics, ConQueR shows great potential in reducing false positives and improving the sparsity of final predictions. We list the average number of predictions per scene for different 3D detectors in Table~\ref{tab:sparsity}. For the baseline Voxel-DETR, thresholding according to scores helps to reduce $\sim$25\% predictions per sample with slightly better performance. With the help of Query Contrast, ConQueR further reduces the number of predictions substantially by $\sim$60\%. Besides, as the performance of ConQueR continually improves (the last two lines), the sparsity of final predictions steadily improve as well. 
When we adopt the same top-300 predictions as baseline Voxel-DETR$_{\rm topN}$ for evaluation, ConQueR$_{\rm topN}$ still improves the detection performance significantly. This indicates the Query Contrast mechanism contributes to generating more accurate predictions from best matched queries.
Furthermore, our ConQueR can achieve much sparser predictions even compared with NMS-based dense detectors such as CenterPoint.

\vspace{1mm}
\subsection{Ablation Study}
\paragraph{Components of Query Contrast.}

We deduce the components of ConQueR to baseline Voxel-DETR by gradually removing multi-positive pairs, auxiliary de-noising loss, and contrastive loss in Table~\ref{tab:QC_components}.  
Compared to ConQueR (the first row), removing the multiple noised copies of GTs from contrastive learning (the second row) causes over 0.6 mAPH/L2 performance drop. If we further remove the auxiliary denoising loss (the third row), performances of vehicles and pedestrians classes even become slightly better, indicating that the auxiliary denoising loss alone is not the key for performance improvements. Moreover, we can find that Query Contrast with only original GTs (the second last entry) already improves over the baseline (the last entry) dramatically especially on pedestrians and cyclists. Overall, the Query Contrast scheme brings \textbf{1.1, 1.7, 2.3} APH/L2 improvements for vehicles, pedestrians and cyclists respectively.

\begin{table}[t]
    \footnotesize
	\begin{center}
		\begin{tabular}{ccc|ccc}
			\toprule
			\multirow{2}{*}{\shortstack[1]{InfoNCE\\Loss}} & \multirow{2}{*}{\shortstack[1]{Aux\\DN}} & \multirow{2}{*}{\shortstack[1]{Multi\\Pos}} & \multicolumn{3}{c}{APH/L2}  \\
			& & & Veh. & Ped. & Cyc.\\
			\midrule
			\checkmark & \checkmark & \checkmark  & \bf68.2  & \bf64.7 & \bf70.1 \\
			\checkmark & \checkmark &  & 67.4 ({\color{red} -0.8}) & 64.1 ({\color{red} -0.6}) & 69.6 ({\color{red} -0.5})   \\
			\checkmark  &  &  & 67.5 ({\color{blue} +0.1}) & 64.2 ({\color{blue} +0.1}) & 69.3 ({\color{red} -0.3}) \\
			   &  &  & 67.1 ({\color{red} -0.4}) & 63.0 ({\color{red} -1.2}) & 67.8 ({\color{red} -1.5}) \\
			\bottomrule
		\end{tabular}
	    \vspace{-4mm}
	\end{center}
	\caption{Effects of components in Query Contrast. The numbers in brackets denotes the performance drop ({\color{red}red}) or increase ({\color{blue}blue}) for each component. Both the multi-positive contrastive loss (Multi-Pos) and the InfoNCE loss (Eq.~(\ref{eq:qc})) from only original GTs have deep impact on performance, while the auxiliary denoising loss (Aux-DN) only has marginal effects.}
	\label{tab:QC_components}
\end{table}

\vspace{-2mm}
\paragraph{Effects of different supervisions or similarity metrics for GT-query pairs.}
We demonstrate the effects of different type of supervision or similarity metrics applied to GT-query pairs in Table~\ref{tab:sim_measure}. As discussed in Sec.~\ref{sec:cns}, simple geometric relations like GIoU cannot sufficiently measure the similarities between GTs and queries because they cannot take the appearance information into account, thus only have marginal effects compared to our baseline Voxel-DETR. If we replace Query Contrast with the MSE loss in knowledge distillation (KD) to supervise positive GT-query pairs, performance of vehicles is still comparable with our Query Contrast strategy (the last entry), but it cannot handle densely populated categories like pedestrians and cyclists, indicating the importance of suppressing negative GT-query pairs in our Query Contrast strategy.

\begin{table}[t]
\footnotesize
\centering
\begin{tabular}{l|c|c|c}
\hline
\multicolumn{1}{c|}{Methods} & Veh. & Ped. & Cyc. \\ \hline
Voxel-DETR       & 67.1 &  63.0 & 67.8 \\ \hline
ConQueR$_{\rm KD-MSE}$     & 68.1 & 63.4 & 68.2 \\ \hline
ConQueR$_{\rm QC-GIoU}$    & 66.6 & 63.6 & 68.4 \\ \hline
ConQueR$_{\rm QC-Cos}$       & \bf68.2 &  \bf64.7 & \bf70.1 \\ \hline
\end{tabular}
\caption{Effects of different supervisions or similarity metrics applied to GT-query pairs. APH/L2 results are reported. $_{\rm QC-Cos}$ denotes our default Query Contrast with the cosine similarity metric, while $_{\rm QC-GIoU}$ denotes using GIoU as the similarity measurement of GT-query pairs. $_{\rm KD-MSE}$ indicates replacing Query Contrast with Knowledge Distillation MSE loss to supervise positive GT-query pairs only.}
\label{tab:sim_measure}
\vspace{-2mm}
\end{table}

\vspace{-2mm}
\paragraph{Number of positive pairs.}
We present the results of using different numbers of noised GT copies in Table~\ref{tab:dn_groups}. We observe that using 3 groups of noised copies without original GTs (default setting) achieves the best performance. Moreover, incorporating original GT into the multi-positive contrastive loss harms the performance. The first two entries show that using single noised copies of GTs is better than using the original GTs. We conjecture this is caused by the lack of training for original GT boxes. The detector is only trained to recover from noised GTs, while having no idea how to deal with perfectly located original GTs.

\begin{table}[t]
\footnotesize
\centering
\begin{tabular}{cc|c|c|c}
\hline
\multicolumn{1}{c|}{Original GTs} & $\#$ Noised GT Groups & Veh & Ped & Cyc             \\ \hline
            \checkmark            & 0             &  67.5 & 64.2 &  69.3        \\ 
\rowcolor[gray]{0.9}
                                  & 1             &  67.9 & 64.4 &  69.6           \\ \cline{1-5} 
                                  & 2             &  68.2 & 64.3 &  69.9        \\ 
\rowcolor[gray]{0.9}
            \checkmark            & 2             &  67.8 & 64.4 &  68.8        \\ \cline{1-5} 
                                  & 3             &  \bf68.2 & \bf64.7 &   \bf70.1        \\ 
\rowcolor[gray]{0.9}
            \checkmark            & 3             & 68.0  & 64.3  &  69.9        \\ 
                                  & 4             &  67.7 & 64.4 &  70.1        \\ \hline
\end{tabular} 
\caption{Number of positive pairs in the contrastive loss. APH/L2 results are reported on the WOD \emph{validation} split. $\checkmark$ denotes including the original GT group into Eq.~(\ref{eq:qc}).}
\label{tab:dn_groups}
\end{table}

\vspace{-2mm}
\paragraph{Query-GT feature alignment.}
We demonstrate the importance of aligning query embeddings to GTs' with an extra MLP in Table~\ref{tab:afp}. Removing the MLP for query embeddings alignment (the first row) or applying the MLP alignment for both GT and query embeddings (the last row) causes $\sim$1 APH/L2 performance drop, indicating the importance of the asymmetric alignment design to mitigate the distribution gap between GT and query embeddings.

\begin{table}[t]
\footnotesize
\centering
\begin{tabular}{c|c|c|c}
\hline
Projection & Veh. & Ped.  & Cyc.     \\ \hline
     & 67.2   &  64.2   &  69.3   \\ \hline
Q    & \bf68.2 & \bf64.7 & \bf70.1 \\ \hline
G\&Q & 67.3    & 64.1    & 68.9   \\ \hline
\end{tabular}
\caption{Design choices of the asymmetric feature alignment. APH/L2 results are reported. `G' and `Q' denotes GT and query embeddings respectively from the selected layer in detector or prediction heads.}
\label{tab:afp}
\vspace{-1mm}
\end{table}

\vspace{-2mm}
\paragraph{Neural Layers for conducting Query Contrast.}
We compare 3 layer alternatives to conduct Query Contrast in Table~\ref{tab:layer_source}: the output layer of each decoder layer, the output layer of each FFN prediction head, and the second-last layer of each FFN prediction head. The Query Contrast scheme can bring consistent improvements for all layer choices, and the features from the last layer of FFN prediction head performs the best, indicating that directly regulate the detection outputs via the contrastive loss can achieve the ``enhance-suppress'' effects onto queries to the utmost.

\begin{table}[t]
\footnotesize
\centering
\begin{tabular}{l|c|c|c}
\hline
Layer to Contrast          & Veh. & Ped. & Cyc.          \\ \hline
Last$_{\rm decoder}$ & 68.1 & 63.9 & 69.7                 \\ \hline
Last$_{\rm FFN}$    &  \bf68.2 & \bf64.7 & \bf70.1        \\ \hline
SecondLast$_{\rm FFN}$ & 67.4 &  64.6 & 69.6                 \\ \hline
\end{tabular}
\caption{Layers to conduct Query Contrast. Results are the APH/L2 reported on the WOD \emph{validation} split. Last$_{\rm decoder}$ and Last$_{\rm FFN}$ denotes the output layer of each decoder layer and FFN prediction head respectively, while SecondLast$_{\rm FFN}$ indicates the second-last layer of each FFN prediction head is chosen to conduct Query Contrast.}
\label{tab:layer_source}
\vspace{-3mm}
\end{table}

\vspace{-3mm}
\paragraph{Generalisation ability w.r.t. query numbers.}
We verify the generalization ability of Query Contrast by varying query numbers in Table~\ref{tab:vary_query_number}. By default we adopt top-1000 scored proposals as initial queries to input to the transformer decoder. The performance gain of Query Contrast is relatively stable when we gradually reduce query numbers to 500 and 300.

\begin{table}[t]
\footnotesize
\centering
\begin{tabular}{l|c|c|c|c}
\hline
\multicolumn{1}{c|}{Methods}  & $\#$Query & Veh. & Ped. & Cyc. \\ \hline
Voxel-DETR & 300       & 66.3 & 62.0 & 66.5 \\ \hline
ConQueR & 300      & 67.0 ({\color{blue} +0.7}) & 63.6 ({\color{blue} +1.6})  & 68.9 ({\color{blue} +2.4}) \\ \hline
Voxel-DETR & 500       & 66.9 & 62.8 & 67.3 \\ \hline
ConQueR  & 500       & 67.8 ({\color{blue} +0.9}) & 64.4 ({\color{blue} +1.6}) & 69.0 ({\color{blue} +1.7}) \\ \hline
Voxel-DETR & 1000      & 67.1 & 63.0 & 67.8 \\ \hline
ConQueR  & 1000      & {\bf68.2} ({\color{blue} +1.1}) & {\bf64.7} ({\color{blue} +1.7}) & {\bf70.1} ({\color{blue} +2.3})  \\ \hline
\end{tabular}
\caption{Improvements of Query Contrast under different query numbers. APH/L2 results are reported. The {\color{blue} blue} numbers in brackets indicates the performance gains.}
\label{tab:vary_query_number}
\end{table}

\vspace{-2mm}
\paragraph{EMA coefficients for generating GT embeddings.}
Here we show results of different momentums of our EMA decoder, which is used to embed GT boxes, in Table~\ref{tab:ema}. The performance of using the same decoder as queries (the first line) already achieves satisfactory results, while introducing a more stable decoder for GT boxes can further improve the performance especially on categories with fewer instances (\emph{i.e.}, cyclists).

\begin{table}[t]
\footnotesize
\centering
\begin{tabular}{c|c|c|c}
\hline
Momentum & Veh. & Ped. & Cyc. \\ \hline
0        & 67.9 & 64.4 & 69.0 \\ \hline
0.9      & 67.6 & 64.3 & 69.1 \\ \hline
0.99     & 68.0 & 64.5 & 69.2 \\ \hline
0.999    & \bf68.2 & \bf64.7 & \bf70.1  \\ \hline
\end{tabular}
\caption{Effects of EMA momentum coefficient.}
\label{tab:ema}
\end{table}

\vspace{-2mm}
\paragraph{Temperature coefficient in Eq.~(\ref{eq:qc}).}
We shown the effects of different $\tau$ in Table~\ref{tab:tau}. $\tau$ controls the contrastive learning difficulty of the GT-query similarities, and we find $\tau=0.7$ leads to the best performance.

\begin{table}[t]
\footnotesize
\centering
\begin{tabular}{c|c|c|c}
\hline 
$\tau$  & Veh.  & Ped.  & Cyc.      \\ \hline
1.0  & 67.9 & 64.2 &  69.8    \\ \cline{1-1} \cline{2-4} 
0.7  & \bf68.2 & \bf64.7 & \bf70.1    \\ \cline{1-1} \cline{2-4} 
0.5  & 67.6 & 64.5 & 69.7     \\ \cline{1-1} \cline{2-4} 
\end{tabular}
\caption{Effects of $\tau$. APH/L2 results are reported.}
\label{tab:tau}
\vspace{-2mm}
\end{table}

\section{Conclusion}

DETR-based sparse 3D detectors faces the problem of duplicated false positives, and lags in detection performance. In this paper, we solve these challenges with our simple yet effective Query Contrast Voxel-DETR (ConQueR). The problem of duplicated false positives is mainly caused by the lack of supervisions in handlings dense similar queries. Based on our sparse 3D detection framework Voxel-DETR, we propose a Query Contrast strategy to explicitly suppress densely overlapping false positives, and simultaneously promote the best matched queries towards their assigned GTs in a contrastive manner. ConQueR reduces $\sim$60$\%$ false positives in the final sparse predictions, closes the gap between sparse and dense 3D detectors, and surpasses previous state-of-the-art 3D detectors by a large margin on the challenging WOD benchmark.

{\small
\bibliographystyle{ieee_fullname}
\bibliography{egbib}
}

\newpage

\appendix

\section{Effects of NMS}
\label{appdenx:a}

Here we demonstrate the effects of NMS on the sparse predictions of our Voxel-DETR and ConQueR in Table~\ref{tab:nms}. Improper NMS configurations (\emph{e.g.}, score and IoU thresholds) can cause performance degradation for all categories. And we find that the NMS configuration adopted by dense detectors (\emph{i.e.}, score threshold 0.1, IoU threshold 0.7) performs the best. For small and densely populated categories such as pedestrians, NMS can bring noticeable performance gains, which can be observed from our baseline Voxel-DETR. However, for the well-trained large vehicles, NMS comes with a significant performance penalty, which in turn demonstrates the effectiveness of our sparse 3D object detection framework. For cyclists, NMS fluctuates in its effects on detection performance, which indicates that NMS is not necessarily required for this category. We conclude that the impact of NMS on detection performance is originates from our baseline Voxel-DETR and the inherent learning difficulty in data for extremely close query predictions, rather the Query Contrast mechanism.

\begin{table}[htbp]
\footnotesize
\centering
\begin{tabular}{l|c|c|c}
\hline
\multicolumn{1}{c|}{Methods}  & Veh.    & Ped.   & Cyc.   \\ \hline
\rowcolor[gray]{0.9}
\multicolumn{4}{l}{\textbf{\emph{validation}} set}\\
Voxel-DETR                    &  68.2 &  64.7 &   70.1 \\ \hline
Voxel-DETR$_{\rm nms}$        &  67.1 ({\color{red}-1.1}) &  67.1 ({\color{blue}+2.4}) &  70.2 ({\color{blue}+0.1}) \\ \hline
ConQueR                       &  70.5 &  68.1 &   73.3  \\ \hline
ConQueR$_{\rm nms}$           &  69.2({\color{red}-1.3}) &  70.1 ({\color{blue}+2.0}) &  74.1 ({\color{blue}+0.8}) \\ \hline
\rowcolor[gray]{0.9}
\multicolumn{4}{l}{\textbf{\emph{test}} set}\\
ConQueR                       & 73.3  & 68.7  & 71.9   \\ \hline
ConQueR$_{\rm nms}$           & 72.6 ({\color{red}-0.7})  & 70.9 ({\color{blue}+2.2})  & 71.7 ({\color{red}-0.2})  \\ \hline
\end{tabular}
\caption{Effects of NMS. APH/L2 results are reported. The numbers in brackets indicates increase ({\color{blue}blue}) or drop ({\color{red}red}) in detection performance. $_{\rm nms}$ denotes further conducting NMS on their corresponding sparse predictions.}
\label{tab:nms}
\end{table}

\end{document}